\documentclass[letterpaper, 10 pt, conference]{ieeeconf}  

\IEEEoverridecommandlockouts                              

\usepackage{amsmath,amsfonts}
\usepackage{algorithmic}
\usepackage{algorithm}
\usepackage{array}
\usepackage[caption=false,font=normalsize,labelfont=sf,textfont=sf]{subfig}
\usepackage{textcomp}
\usepackage{stfloats}
\usepackage{url}
\usepackage{verbatim}
\usepackage{graphicx}
\usepackage{cite}
\usepackage{booktabs}
\usepackage{multirow} 
\usepackage{makecell}
\usepackage{tabularx}   
\usepackage{xcolor,soul}
\usepackage[normalem]{ulem}
\title{\LARGE \bf
Modeling and Control of a Pneumatic Soft Robotic Catheter Using Neural Koopman Operators
}

\author{Yiyao Yue$^{1}$, Noah Barnes$^{2}$, Lingyun Di$^{1}$, Olivia Young$^{3}$, Ryan D. Sochol$^{3}$, \\ Jeremy D. Brown$^{2}$ and Axel Krieger$^{2}$
\thanks{*This work was supported by the National Institutes of Health (R01EB033354), 
the Maryland Robotics Center and the Center for Engineering Concepts 
Development at the University of Maryland, and the National Science 
Foundation Graduate Research Fellowship Program (DGE 2236417, 2139757). 
Any opinions, findings, and conclusions expressed are those of the authors 
and do not necessarily reflect the views of the National Science Foundation.}
\thanks{$^{1}$Laboratory for Computational Sensing and Robotics, Johns Hopkins
 University,
        Baltimore, MD, USA.
         Email: \{yyue19, ldi4\}@jh.edu}%
\thanks{$^{2}$Department of Mechanical Engineering, Johns Hopkins University, 
        Baltimore, MD, USA.
         Email: \{nbarne18, jdelainebrown, axel\}@jhu.edu}%
\thanks{$^{3}$Department of Mechanical Engineering, University of Maryland, 
        College Park, MD, USA.
         Email: \{oyoung, rsochol\}@umd.edu}%
}

\begin{document}

\maketitle
\thispagestyle{empty}
\pagestyle{empty}

\begin{abstract}

Catheter-based interventions are widely used for the diagnosis and treatment of cardiac diseases. Recently, robotic catheters have attracted attention for their ability to improve precision and stability over conventional manual approaches. However, accurate modeling and control of soft robotic catheters remain challenging due to their complex, nonlinear behavior. The Koopman operator enables lifting the original system data into a linear ``lifted space'', offering a data-driven framework for predictive control; however, manually chosen basis functions in the lifted space often oversimplify system behaviors and degrade control performance. To address this, we propose a neural network-enhanced Koopman operator framework that jointly learns the lifted space representation and Koopman operator in an end-to-end manner. Moreover, motivated by the need to minimize radiation exposure during X-ray fluoroscopy in cardiac ablation, we investigate open-loop control strategies using neural Koopman operators to reliably reach target poses without continuous imaging feedback. The proposed method is validated in two experimental scenarios: interactive position control and a simulated cardiac ablation task using an atrium-like cavity. Our approach achieves average errors of $2.1 \pm 0.4$ mm in position and $4.9 \pm 0.6^\circ$ in orientation, outperforming not only model-based baselines but also other Koopman variants in targeting accuracy and efficiency. These results highlight the potential of the proposed framework for advancing soft robotic catheter systems and improving catheter-based interventions.

\end{abstract}


\section{Introduction}

\begin{figure}[!t]
    \centering
    \centerline{\includegraphics[width= 1 \columnwidth]{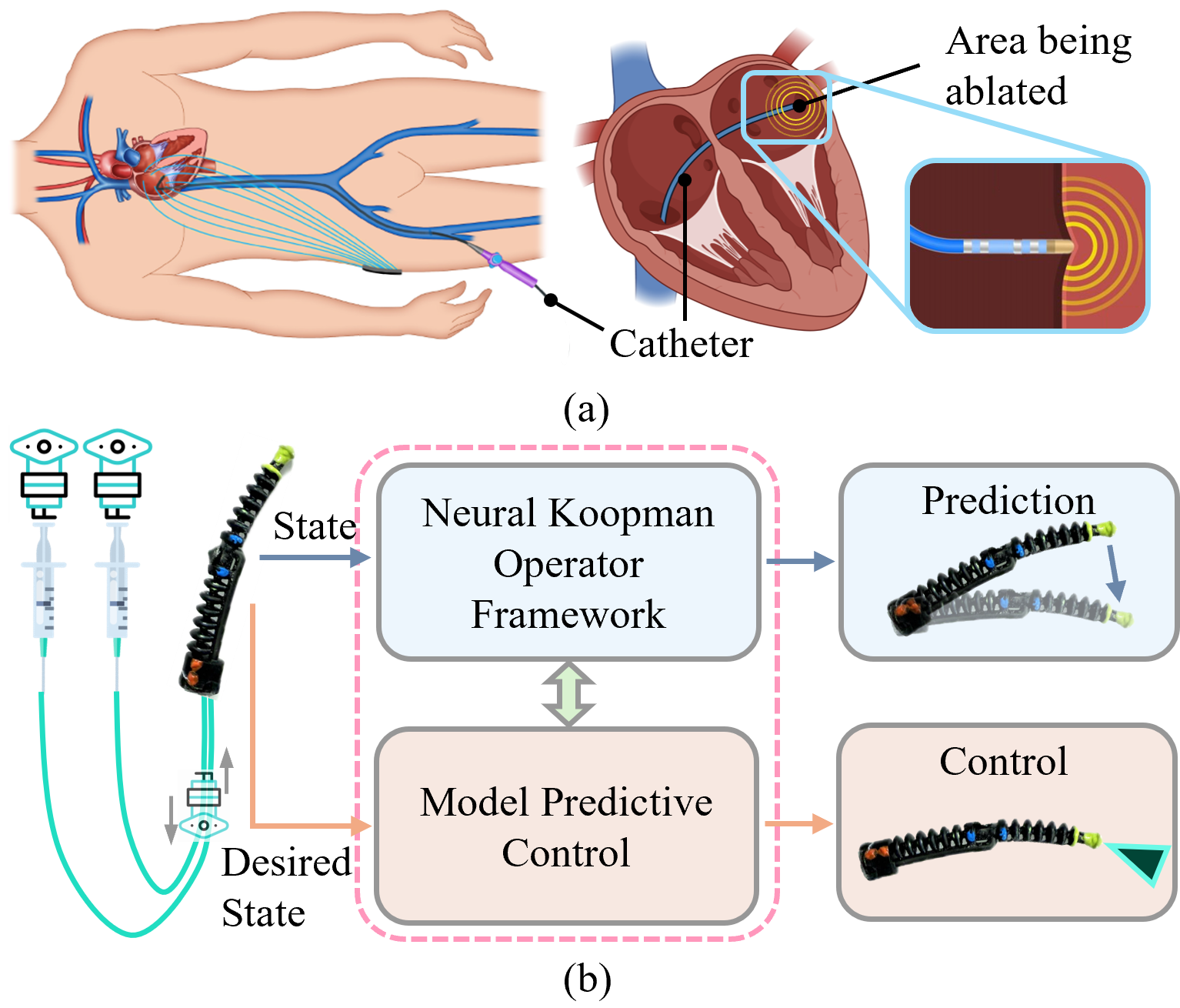}}
    \caption{Overview of soft robotic catheter applications for atrium ablation and control framework. (a) Illustrative robotic catheter ablation for atrial fibrillation. (b) Neural Koopman operator framework with model predictive control method.}
    \label{abstract}
\end{figure}

Cardiac catheter interventions, require precise navigation through tortuous and fragile vasculature, making them highly dependent on operator skill and manual dexterity. Robotic catheter systems have emerged as a promising solution to improve maneuverability, precision, and safety in delicate cardiovascular environments \cite{konda2025robotically}.
One important application of cardiac catheter interventions is to treat abnormal heart rhythms caused by atrial fibrillation \cite{wijesurendra2019mechanisms}. This procedure requires the robotic catheter to precisely navigate and target the lesion site \cite{bassil2020robotics}. The illustrative robotic catheter application for atrium ablation is shown in Fig.~\ref{abstract}(a).

In clinical practice, catheter navigation is constrained by limited and intermittent sensing. Imaging modalities such as fluoroscopy provide sparse guidance and need to be minimized due to radiation exposure risks \cite{ferguson2009catheter}. Although zero-fluoroscopy workflows have been demonstrated \cite{enriquez2024feasibility, prolivc2022conventional}, continuous feedback remains difficult to obtain. As a result, navigating a catheter through tortuous and fragile vasculature without real-time imaging poses significant challenges for maintaining motion accuracy and safety. Furthermore, embedding dense sensors in soft robotic catheter is challenging due to miniaturization and compliance requirements. These sensing limitations motivate the development of open-loop control strategies, where catheter motion must remain predictable and reliable without continuous feedback.

Achieving reliable performance requires accurate modeling and control of pneumatic soft robotic catheters (PSRCs), which are inherently underactuated and highly nonlinear. 
Traditional model-based approaches, including the piecewise constant curvature (PCC) model \cite{webster2010design}, rely on simplifying geometric assumptions that limit generalization to pneumatic systems. Physics-based formulations such as the Cosserat rod model \cite{shi2024position} require accurate boundary conditions and material parameters, making them sensitive to manufacturing inconsistencies. Finite-element methods (FEMs) \cite{ferrentino2023finite} capture rich dynamics but are computationally prohibitive for real-time control.

Learning-based approaches provide an alternative when analytical modeling is intractable \cite{laschi2023learning}. Methods such as learning the inverse kinematics \cite{hao2023inverse}, motion compensation \cite{yao2025adaptive}, or Jacobian estimation \cite{fang2022efficient} have shown improved accuracy. However, these approaches are difficult to integrate with existing model-based control frameworks, limiting their applicability in practice \cite{bruder2020data}.

Recently, the Koopman operator has shown strong potential for data-driven modeling and control of soft robots by ``lifting'' the system dynamics into a higher-dimensional space where the nonlinear evolution can be approximated linearly \cite{bruder2020data}. EDMD-based approaches for concentric tube robots follow a similar lifting principle for predictive control \cite{thamo2022data}. However, the effectiveness of these methods strongly depends on the choice of basis functions, which are often manually designed and can limit robustness and generalization \cite{han2023robust}. Neural networks have been introduced to automate the lifting process, with applications extending beyond soft robotics \cite{xiao2022deep,han2023robust}. However, many existing frameworks assume assume that the input affects the system linearly, restricting their ability to capture complex nonlinear behaviors. In parallel, purely neural dynamic models have been explored for soft robot control \cite{thuruthel2017learning,gillespie2018learning}, where the network directly approximates the nonlinear dynamics in the original state space. Although expressive, do not introduce a structured transformation of the dynamics into a higher-dimensional linear representation, which may limit their interpretability of the learned dynamics. Koopman–neural network methods have also been used for the spectral analysis of robot dynamics \cite{komeno2022deep}. However, this work focused on an externally actuated robotic arm without providing a suitable way to drive the PSRC to a target.

Despite recent progress, two key challenges remain in Koopman-based PSRC modeling and control. First, lifting functions are often designed manually, limiting the adaptability to complex nonlinear behaviors. Second, achieving accurate open-loop control is difficult in catheter-based applications, where target reaching must be reliable without continuous feedback. To address these challenges, we propose a neural Koopman operator framework that models the nonlinear behavior of the PSRC and enables open-loop control via model predictive control (MPC) in the lifted space (Fig.~\ref{abstract}(b)). The framework jointly learns state and input encoders, the Koopman operator, and corresponding decoders from data. We validate the approach through model prediction and two real-world experiments.


The main contributions are as follows: (1) We propose a data-driven framework that integrates Koopman operator theory and neural networks, preserving a linear structure in the lifted space while capturing nonlinear PSRC behaviors. (2) We integrate the framework with MPC to achieve accurate open-loop control. (3) We evaluate both modeling and control performance in position and pose regulation tasks, including a clinically inspired scenario, demonstrating improved accuracy and efficiency over baseline methods.

\section{System Modeling}
\subsection{Koopman Operator Background}
In our problem setting, the PSRC is actuated by two inputs $u = [u_1, u_2]^T$, and the system state $x = [x, y, \theta]^T$ represents the tip position and orientation. To address the challenge of modeling such nonlinear behavior, we adopt the Koopman operator theory. The following summarizes the derivation of the Koopman operator, which we will later extend in our neural Koopman framework for PSRC modeling.

Consider a dynamical nonlinear system:
\begin{equation} \label{eq:sample}
x_{k+1} = F(x_k, u_k),
\end{equation}
where $x \in \mathbb{R}^n$ is the system state, $u \in \mathbb{R}^m$ as system control input.
The nonlinear system can be lifted to an infinite-dimensional function space \cite{budivsic2012applied} in which the system becomes linear, where the evolution of the system can be characterized by the Koopman operator $\mathcal{K}$:
\begin{equation} \label{eq:sample}
\varphi(x_{k+1})=\mathcal{K}\varphi(x_k),
\end{equation}
where $\varphi(x_k)$  is the observable in the lifted space. Then we define a set of observables: 
\begin{equation} \label{eq:sample}
\varphi(x_k) = [\varphi_1(x_k)^\top,...,\varphi_N(x_k)^\top]^\top,
\end{equation}
where  $\varphi(x_k) \in \mathbb{R}^N$. The components of $\varphi(x_k)$ are basis functions manually chosen in the lifted space. There are several possible basis functions including, such as monomial basis \cite{bruder2020data} and radial basis functions (RBF) \cite{xiao2022deep}. Then, considering the assumption that $u$ is constant, the system evolution in the lifted space can be represented as:

\begin{equation} \label{eq:sample}
\varphi(x_{k+1})  = 
\begin{bmatrix}
A&B
\end{bmatrix}
\begin{bmatrix}
\varphi(x_{k}) \\
u_{k}
\end{bmatrix},
\end{equation}
where $A$ and $B$ are embedded in $\mathcal{K}$ with the form:

\begin{equation} \label{eq:sample}
\mathcal{K}=
\begin{bmatrix}
A_{N \times N} & B_{N \times m}\\
O_{m \times N} & I_{m \times m}
\end{bmatrix}.
\end{equation}

The best $A$ and $B$ matrices can be isolated after finding the best $\mathcal{K}$ by solving
\begin{equation} \label{eq:sample}
\mathcal{K}=G^{\dagger}H,
\end{equation}
where
\begin{equation} \label{eq:sample}
G=
\begin{bmatrix}
\varphi(x_1)&...& \varphi(x_K) \\
u_1 &...& u_K
\end{bmatrix}^\top,
H=
\begin{bmatrix}
\varphi(x_1^+)&...& \varphi(x_K^+) \\
u_1 &...& u_K
\end{bmatrix}^\top
\end{equation}
and $x_i^+$ is the evolution result of $x_i$ and $u_i$, $K$ is the total number of state measurements. We can define the projection matrix $C \in \mathbb{R}^{n \times N}$ which satisfies $x_{k+1}=C\varphi(x_{k+1})$. By sorting the order of the basis in (3), the $C$ can be represented as:
\begin{equation} \label{eq:sample}
C=
\begin{bmatrix}
O_{n \times n} & I_{n \times (N-n))}
\end{bmatrix}.
\end{equation}
 A more detailed derivation can be found in \cite{mauroy2019koopman} and \cite{bruder2020data}.
 
However, such a standard approach and approximation offer limited prediction and control performance. In particular, choosing basis functions without prior knowledge, especially for PSRC, may fail to capture essential behaviors. These limitations motivate learning more flexible lifting representations with neural Koopman operator frameworks.

\subsection{Neural Koopman Operator Framework}


\begin{figure*}[!t]
    \vspace{2mm}  
    \centering
    \includegraphics[width=0.98\linewidth]{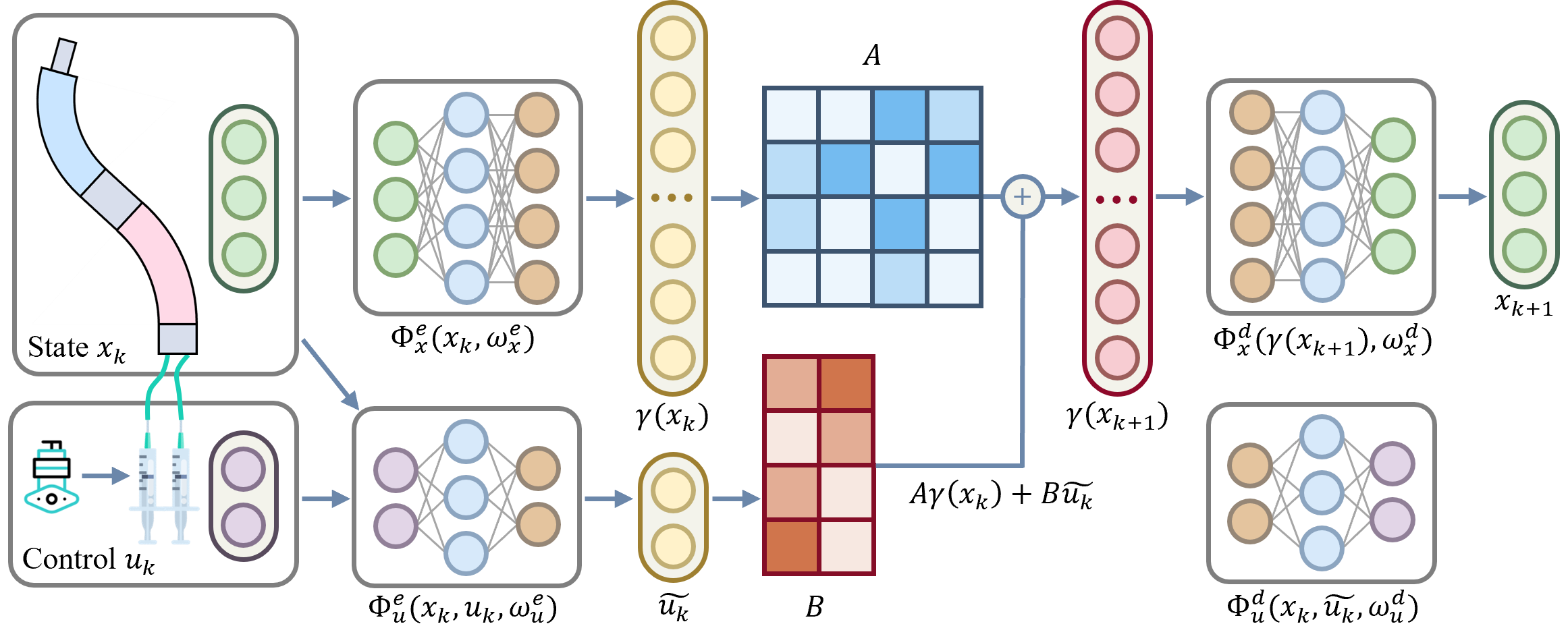}
    \caption{The diagram of neural Koopman operator framework. The original state $x_k$ is lifted by encoder  $\Phi_x^e(x_k, \omega_x^e)$. And the $\tilde{u_k}$ is encoded by $\Phi_u^e(x_k, u_k,\omega_u^e)$ to construct the evolution in lifted space. Original state and input can be recovered by decoders $\Phi_x^d(\gamma(x_{k+1}), \omega_x^d)$ and $\Phi_u^d(x_k,\tilde{u_k}, \omega_u^d)$.}
    \label{framework}
\end{figure*}

The neural Koopman operator framework employs a feedforward neural network (FNN) to construct a complete lifting dictionary and a decoder to map the observables in the lifted space back to the original states. In the context of our PSRC, this framework takes the input pressures as control inputs, and the robot tip pose or position as the system states. The overall proposed framework is illustrated in Fig.~\ref{framework}. According to equation (4), the system evolution in the lifted space of the neural Koopman operator framework can be represented as:
\begin{equation} \label{eq:sample}
\gamma(x_{k+1}) = A\Phi_x^e(x_k,\omega_x^e) + B\Phi_u^e(x_k, u_k, \omega_u^e),
\end{equation}
where $\Phi(\cdot,\omega)$ is a neural network parameterized with weights $\omega$, $\gamma(x_{k+1}) \in \mathbb{R}^N$, $\Phi_x^e(x_k,\omega_x^e) \in \mathbb{R}^N$ and $\Phi_u^e(x_k, u_k, \omega_u^e) \in \mathbb{R}^m$. The encoder network $\Phi_x^e(x_k,\omega_x^e)$ is designed to map the original state into the lifted space. Instead of concatenating the original state with the encoder output \cite{shi2022deep} to simplify the decoder mapping matrix, this paper directly encodes the entire original state and introduces a separate decoder network.

The next part of the framework is the input encoder network $\Phi_u^e(x_k, u_k, \omega_u^e)$. We first propose a version of this structure that encodes both $x_k$ and $u_k$ together to form a nonlinear input neural Koopman (NINK). The second version, called LINK, takes a linear input representation; thus, the encoder network $\Phi_u^e(x_k, u_k, \omega_u^e)$ degrades to just $u_k$. The system evolution of the LINK structure is then expressed as:
\begin{equation} \label{eq:sample}
\gamma(x_{k+1}) = A\Phi_x^e(x_k,\omega_x^e) + Bu_k.
\end{equation}

For simplicity, we use the notation $\tilde{u}$ to denote the output of the input encoder network in both NINK and LINK.

By training the entire network end-to-end, the proposed structure enables joint training of the encoders and the system matrices $A$ and $B$, ensuring the best approximation of the system evolution.

Similarly, the decoder network $\Phi_x^d(\gamma(x_k),\omega_x^d)$ that maps the lifted space $\gamma(x_k)$ back to the original state $x_k$ should satisfy:
\begin{equation} \label{eq:sample}
x_{k} = \Phi_x^d(\gamma(x_k),\omega_x^d).
\end{equation}

Next, with the neural Koopman operator framework established, we now describe the training procedure, which consists of two main components: the encoder networks and the decoder network.

The main objective of learning the encoder network is to find a set of parameters ($\omega_x^e$, $\omega_u^e$, $A$, $B$ for NINK; $\omega_x^e$, $A$, $B$ for LINK) that minimize the prediction error by:
\begin{align} \label{eq:sample}
L(\omega, A,B)&=\frac{1}{R} \sum_{i=1}^R \| \gamma_i - \hat{\gamma}_i \|_2^2\\
=\frac{1}{R} \sum_{i=1}^R \|&A\Phi_x^e(x_i,\omega_x^e) - [A\Phi_x^e(x_{i-1},\omega_x^e)+B\widetilde{u_{i}}]\|_2^2,
\end{align}
where $R$ denotes the prediction horizon, $\gamma_i$ is encoded from the $i^{\text{th}}$ state $x_i$, and $\hat{\gamma}_i$ is the predicted value based on the previous state $x_{i-1}$ and control input $u_{i-1}$.

After training the encoder network, the decoder network is optimized by learning parameters $\omega_x^d$ that minimize the reconstruction error, defined as:
\begin{equation} \label{eq:sample}
L(\omega_x^d)= \frac{1}{R} \sum_{i=1}^R \| x_i - \hat{x}_i \|_2^2=\frac{1}{R} \sum_{i=1}^R \| x_i - \Phi_x^d(\gamma(x_i),\omega_x^d) \|_2^2.
\end{equation}

To further enhance long-horizon accuracy, an $R$-step loss is adopted, which minimizes the multi-step prediction errors rather than only the one-step error, thereby encouraging the model to capture long term system evolutions \cite{han2023robust}.



\section{Control Strategies}
In this section, we present two control strategies: a Koopman-based MPC approach and a PSRC-specific PCC Jacobian-based control approach. The latter is employed as a baseline for comparison.

\subsection{Model Predictive Control}
After establishing the neural Koopman operator framework, the control inputs can be optimized via a quadratic loss function of a model-based controller. Specifically, the MPC method makes use of the system model to predict the future behavior of the controlled system and determines the next step's control action \cite{schwenzer2021review}. 

\begin{algorithm}[!ht]
\caption{Neural Koopman MPC (Open-loop)}
\label{alg:koopman_mpc}
\begin{algorithmic}[1]   
\STATE \textbf{Input:} 
\begin{tabular}[t]{@{}l}
  Koopman matrices $A,B$, encoders $\Phi_x^e,\Phi_u^e$, \\
  decoders $\Phi_x^d,\Phi_u^d$ initial state $x_0$, \\
  desired target $x_i^{des}$, control steps $K$, \\
  cost matrices $Q,R$, horizon $R$
\end{tabular}
\FOR{$k = 0,1,\dots,K-1$}
    \STATE Lifting: $\gamma(x_k)=\Phi_x^e(x_k, \omega_x^e)$
    \STATE Solve MPC to obtain $\{\tilde{u}_i^*\}_{i=0}^{R-1}$
    \STATE Recover optimal input: $u_k^* = \Phi_u^d(x_k, \tilde{u}_{i=0}^*)$
    \STATE Predict next lifted state: $\gamma(x_{k+1})$ via (9)
    \STATE Recover next state: $x_{k+1}$ via (11)
    \STATE Apply $u_k^*$ to the PSRC
\ENDFOR
\end{algorithmic}
\end{algorithm}

Since the Koopman operator transforms the nonlinear evolution into a linear lifted space, the optimal control problem is formulated in this space with the control variable $\Phi_u^e$. The resulting MPC problem is expressed as:
\begin{align} \label{eq:sample}
\min_{\tilde{u_i}} \; &(\gamma(x_R)-\gamma(x_R^{des}))^\top Q (\gamma(x_R)-\gamma(x_R^{des}))+\nonumber \\
+&\sum_{i=0}^{R-1} (\gamma(x_i)-\gamma(x_i^{des}))^\top Q (\gamma(x_i)-\gamma(x_i^{des}))+\tilde{u_i}^{\top}R\tilde{u_i}\nonumber\\
&\text{s.t.}\; \; \gamma(x_{i+1})=Ax_i+B\tilde{u_i}, \; \; \gamma(x_0)=\Phi_x^e(x_0,\omega_x^e)\nonumber\\
&\tilde{u_i}=\Phi_u^e(\Phi_x^d(x_i),u_i,\omega_u^e),
\end{align}
where $Q\in \mathbb{R}^{N \times N}$ and $R\in \mathbb{R}^{m \times m}$ are positive semi definite matrices, $R$ is the prediction horizon. For NINK, the MPC yields the optimal encoded input $\tilde{u}_i^*$, which cannot be directly applied to the PSRC. To obtain the executable input, a decoder network is employed:
\begin{equation} \label{eq:sample}
u_i^* = \Phi_u^{d}(x_i,\tilde{u}_i^*, \omega_u^d).
\end{equation}
The open-loop procedure of the neural Koopman MPC controller is summarized in Algorithm~\ref{alg:koopman_mpc}. We evaluate four control methods: NINK + MPC (NNKM), LINK + MPC (LNKM), Monomial basis Koopman operator + MPC (MBKM), state space + MPC (SSM). The MBKM baseline follows prior Koopman-based soft robot control studies \mbox{\cite{bruder2020data}}, adopting a monomial lifting strategy. The SSM baseline is a linear state-space model identified directly in the original state space, without lifting.

\subsection{Piecewise Constant Curvature Control for PSRC}

\begin{figure}[!t]
    \vspace{2mm}
    \centering
    \centerline{\includegraphics[width= 0.96 \columnwidth]{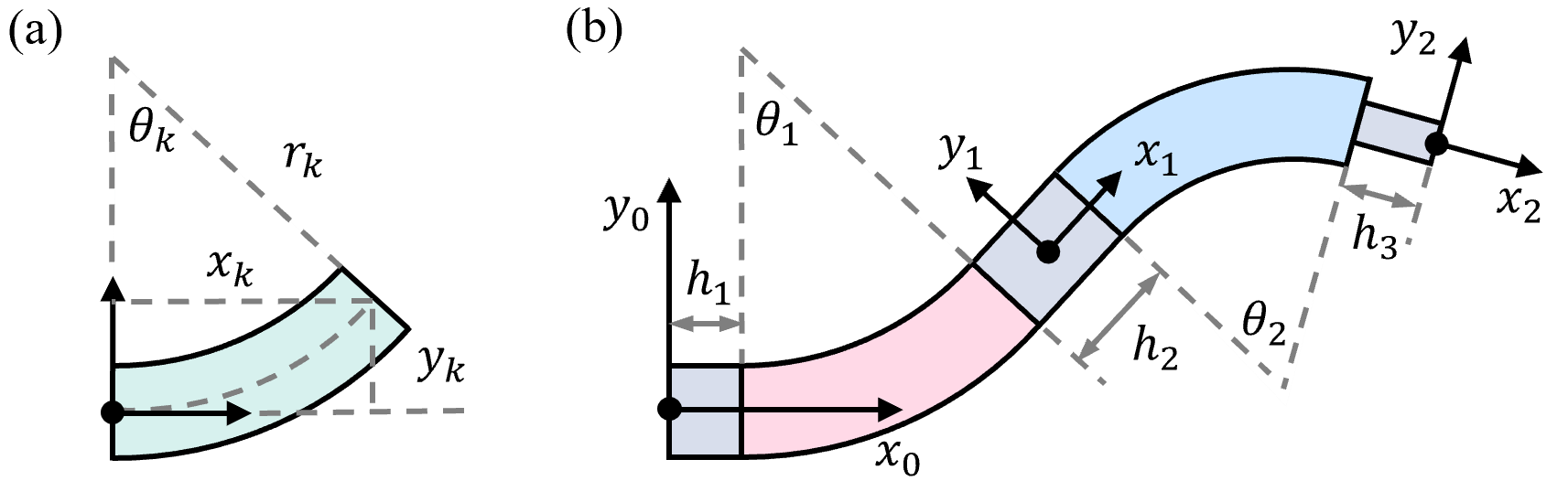}}
    \caption{Geometry assignment of the PCC model. (a) Single segment. (b) Dual segment of the PSRC.}
    \label{pcc}
\end{figure}

To establish a baseline model-based control method for the PSRC, we adopt the widely used PCC assumption \cite{webster2010design}. The PCC model enables an analytical representation of the robot configuration, facilitating Jacobian-based control. Fig.~\ref{pcc}(a) shows the geometry of a single bending segment, and accordingly, the tip position can be expressed as:
\begin{equation} \label{eq:sample}
\begin{bmatrix}
x_k\\y_k
\end{bmatrix}=
\begin{bmatrix}
\frac{l_k}{\theta_k}\sin(\theta_k) \\
\frac{l_k}{\theta_k}\cos(1-\theta_k)
\end{bmatrix},
\end{equation}
where $\l_i$ is the length of the segment. The rigid transformation from the base to the tip can be expressed differently for a bending segment and a straight segment (see Fig.~\ref{pcc}(b)), respectively: 

\noindent
\begin{minipage}{0.53\linewidth}
\begin{equation}\label{eq:Tbend}
{}^{b}T_k=
\begin{bmatrix}
R_z(\theta_k)& 
\begin{bmatrix}
x_k\\y_k
\end{bmatrix}
\\
0&1
\end{bmatrix}
\end{equation}
\end{minipage}\hfill
\begin{minipage}{0.43\linewidth}
\begin{equation}\label{eq:Tstraight}
{}^{s}T_k=
\begin{bmatrix}
I& 
\begin{bmatrix}
0\\h_k
\end{bmatrix}
\\
0&1
\end{bmatrix}
\end{equation}
\end{minipage}
Then, the overall transformation from the robot base to the tip is given by:
\begin{equation}
T = {}^{s}T_1{}^{b}T_1{}^{s}T_2{}^{b}T_2{}^{s}T_3.
\end{equation}
The relationship between actuation space velocity and task space velocity can be written as:
\begin{equation}
\dot{x} = J \dot{q} = \frac{\partial x}{\partial \tilde{\theta_j}}\frac{\partial \tilde{\theta_j}}{\partial q}\dot{q},
\end{equation}
where $\tilde{\theta_j} = [\theta_1, \theta_2, l_1, l_2]$ denotes the joint space parameters, and $q = [q_1, q_2]$ represents the actuation inputs \cite{barnes2024bedside}. In this baseline model, we model the mappings from the actuation space to the joint space $\theta_1=f_1(q_1)$, $l_1=f_2(q_1)$, $\theta_2=f_3(q_2)$, and $l_2=f_4(q_2)$ via second-order polynomial fitting. In contrast to tendon-driven systems \cite{webster2010design}, pneumatic actuation lacks an explicit analytical model, requiring empirical approximation. With these relations, a resolved rate controller can be obtained \cite{lai2021verticalized}:

\begin{equation}
\Delta q = J^{\dagger}P\Delta x_{des},
\end{equation}
where $P$ is a symmetric positive definite matrix. To address the redundancy of the system and avoid infinitely many solutions, we adopt the damped least squares (DLS) pseudo-inverse approach to get the optimized control input \cite{luo2024efficient}:
\begin{equation} \label{eq:sample}
\operatorname*{argmin}_{\Delta q} (\| J\Delta q- P\Delta x_{des} \|_2^2+\lambda\|\Delta q\|_2^2),
\end{equation}
where $\lambda$ represents a positive damping constant. To serve as a baseline model, the PSRC-specific PCC Jacobian-based control is first tested offline; then, the resulting control sequence is applied to the PSRC still in an open-loop manner.


\section{Experiments and Results}

\subsection{Experimental Setup}
\begin{figure}[!t]
    \vspace{2mm}  
    \centering
     \includegraphics[width=0.97\linewidth]{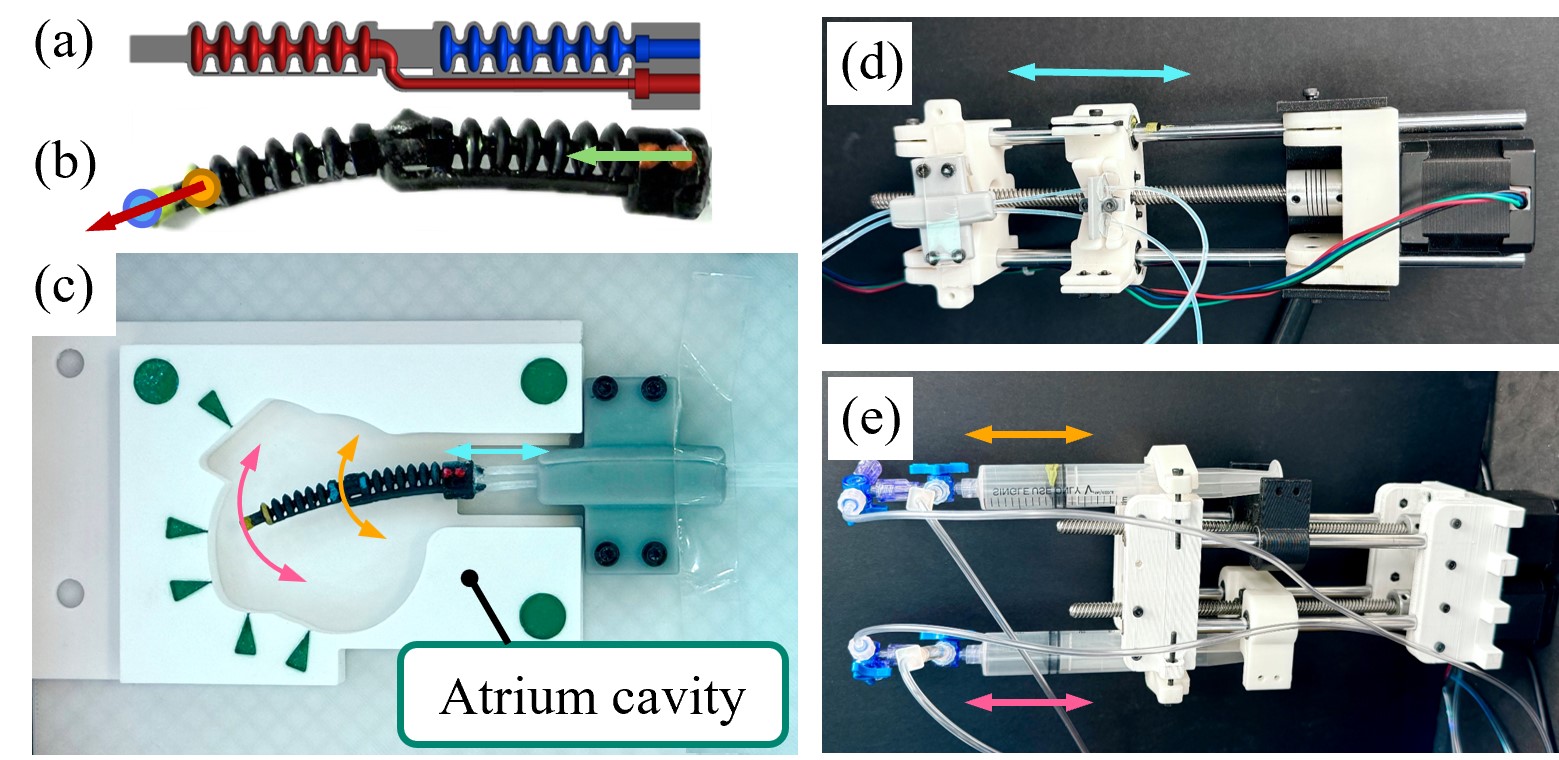}
    \caption{Experimental setup. (a) CAD design of the PSRC. (b) Fabricated PSRC indicating tip vector and base vector. (c)  PSRC positioned inside an atrium model, with an overhead camera view. Colored arrows represent two bending directions and one translation direction. (d) Translation stage for precise positioning. (e) Dual syringe pumps for pressure control of bending.}
    \label{exp}
\end{figure}

In this work, a dual-segment steerable soft robotic catheter was fabricated using additive manufacturing with hyperelastic material (Mars 3, Elegoo, China). The dual-channel design and fabricated prototype are shown in Fig.~\ref{exp}(a)-(b). The tip position and orientation are extracted from color markers. Detailed design and characterization are reported in \cite{jiang2024one}. As shown in Fig.~\ref{exp}(c), experiments were conducted in a custom chamber with an overhead camera (Realsense, Intel, USA) for marker tracking. A 3D-printed atrium-like cavity with predefined pose targets was used for validation. The robot’s steering motion is driven by a NEMA 17 stepper motor (Rtelligent, China), while bending is actuated by two motor-driven syringe pumps with pressure feedback from sensors (MPRLS, Adafruit, USA). All actuators and sensors are controlled via an Arduino Uno board(Arduino, Italy).

\subsection{Data Collection and Model Identification}
We construct a compact dataset of 2,586 input–output samples, where the inputs are the two chamber pressures and the outputs are the robotic tip state (2D position or pose). At step $k$, each syringe chamber’s pressure target is generated by a random increment or decrement relative to its previous target:
\begin{equation} \label{eq:sample}
q_{k+1}=q_k+d_k\Delta q_{k,k+1},
\end{equation}
where $d_k \in \{-1,1\}$ indicates the direction of change, and $\Delta q_{k,k+1}\in\{0,1,\dots,10\}$ kPa denotes the magnitude of the pressure difference between consecutive targets. Both $d_k$ and $\Delta q_{k,k+1}$ are chosen randomly. The pressure targets are tracked using motor-driven syringe pumps under PID control with sensor feedback. To ensure that the commanded pressure reaches its target within 1.5 s, the motor velocity is scaled proportionally to $\Delta q_{k,k+1}$. The data collection was divided into five trials, with 500, 500, 500, 500, and 586 samples, respectively. The robotic tip state is detected by the camera at 2 Hz, and the data collection takes less than 6 minutes per trial.

For all four FNNs in the neural Koopman framework ($\Phi_e$, $\Phi_u$, $\Phi_d$, $\Phi_u^{-1}$), we used a batch size of 8, learning rate of 0.001, and a maximum of 200 epochs. The state encoder $\Phi_e$ outputs a lifted state of dimension $N=30$, with 2 hidden layers of 128 neurons each, which is the same for the decoder $\Phi_d$. The input encoder $\Phi_u$ and its inverse $\Phi_u^{-1}$ both have 2 hidden layers of 32 neurons each. Training was performed in PyTorch on an NVIDIA GeForce RTX4070 Laptop GPU and took 840.2 s. For comparison, the MBKM uses $\omega=2$, yielding $N={(n+m+\omega)!}/{((n+m)!\omega!)}=35$ basis functions for 2D pose ($m=3$) \mbox{\cite{bruder2020data}}, selected to match the neural Koopman lifted dimension ($N=30$).



\subsection{Modeling Performance Evaluation}
\begin{figure}[!t]
    \vspace{2mm}  
    \centering
    \includegraphics[width=0.97\linewidth]{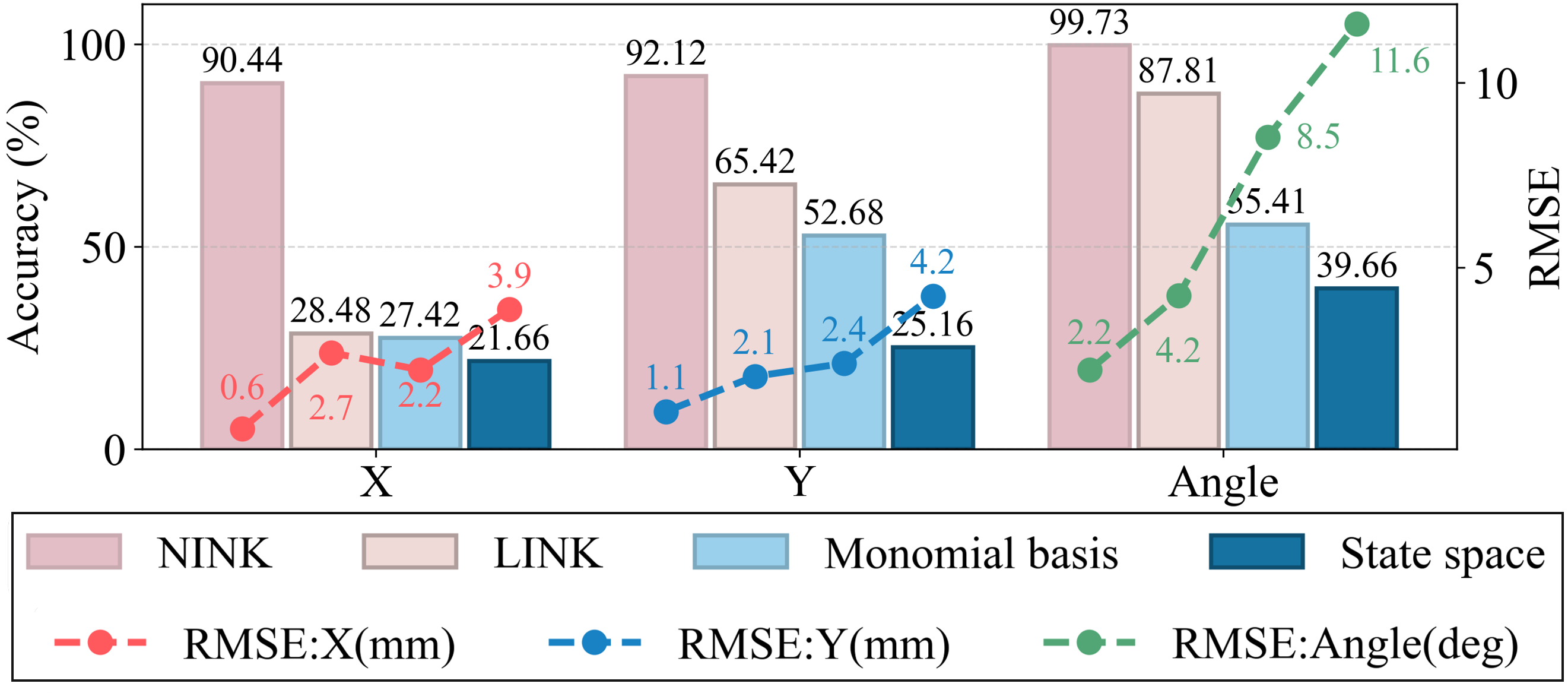}
    \caption{Modeling performance evaluated by prediction accuracy and RMSE across different models. The first row of the legend represents four models. The second row represents RMSE across x and y positions and orientations. }
    \label{model_val}
\end{figure}

In this experiment, four models are trained from the collected data: NINK and LINK models with 30 lifted space dimensions described in Section II-B; a monomial basis Koopman operator model with degree $\omega=2$; and a linear state-space model \cite{della2023model} as a baseline. There are three input states, indicating a 2D pose. Following the training phase, we evaluate the models on unseen data. Two evaluation metrics are defined as follows: Root Mean Square Error (RMSE):
\begin{align}
\text{RMSE} = \sqrt{\frac{1}{N} \sum_{k=1}^{N} (x_k - \hat{x}_k)^2},  \;\;\;k=1,2,...,N,
\end{align}
Accuracy within threshold value (Acc):
\begin{align}
\text{Acc} = \frac{1}{N} \sum_{k=1}^{N} \mathbf{1}\{|x_k - \hat{x}_k|\leq\sigma\},  \;\;\;k=1,2,...,N,
\end{align}
where $N=500$, $\hat{x}_k$ is the predicted value of the states, and $\sigma$ is the threshold value selected as $5.00\%$ of the states' range. The threshold was selected based on the practical positioning tolerance observed in the experimental setup, following a consistent selection principle across all experiments. Note that all orientation angles are within $(-90^\circ,90^\circ)$, which ensures that direct subtraction is valid without angle wrapping.

The modeling performance results are presented in Fig.~\ref{model_val}. Among the four models, the NINK model achieves the best prediction performance across all three states in terms of both Acc (x: 90.44\%, y: 92.12\%, angle: 99.73\%) and RMSE (x: 0.6mm, y: 1.1mm, angle: 2.2mm). The LINK model ranks second, benefiting from the FNN structure, which provides a stronger capability to learn lifted-space representations compared to the manually selected monomial basis in the classical Koopman operator method. However, LINK is limited by the assumption that the control input affects the system linearly \cite{bruder2019nonlinear}. Under pressure-based actuation, this assumption is violated, whereas the input encoder network in NINK offers a more flexible representation of input influence on the system. Notably, while LINK, the monomial basis Koopman operator, and the linear state-space model all exhibit poor accuracy in predicting the $x$ position state, NINK yields substantially better results. This discrepancy arises because, compared with the $y$ position, the $x$ position has a much narrower range in the 2D workspace, making it more challenging to predict. A similar phenomenon has been reported for the $z$ position in 3D single-segment pneumatic soft robots \cite{bruder2019nonlinear}. Finally, the linear state-space model suffers from its linearity assumption, making it unsuitable for modeling and control of pneumatically driven soft robots.

In the following sections, we evaluate the proposed modeling and control framework through two tasks: interactive position control and pose control in an atrium model.

\subsection{Experiment 1: Interactive Position Control}

\begin{table}[t]
\vspace{2mm} 
\centering
\caption{Results of Experiment 1: Euclidean distance error statistics (unit: mm). Two accuracy values are reported with thresholds $\sigma_1=1.5$\,mm and $\sigma_2=3.5$\,mm.}
\label{tab:error_stats}
\begin{tabular}{lccccc}
\toprule
\textbf{Method} & \textbf{AVG} & \textbf{STD} & \textbf{MAX} & \textbf{Acc(1.5)} & \textbf{Acc(3.5)} \\
\midrule
NNKM & \textbf{0.7} & \textbf{0.4} & \textbf{1.2} & \textbf{100.00\%} & \textbf{100.00\%} \\
LNKM & 2.0 & 0.9 & 3.1 & 50.00\% & 83.33\% \\
MBKM & 4.1 & 1.3 & 6.6 & 0.00\% & 16.67\% \\
SSM  & 8.0 & 2.8 & 11.9 & 0.00\% & 0.00\% \\
\bottomrule
\end{tabular}
\end{table}

\begin{figure*}[!t]
\centering
\subfloat{\includegraphics[width=1.75in]{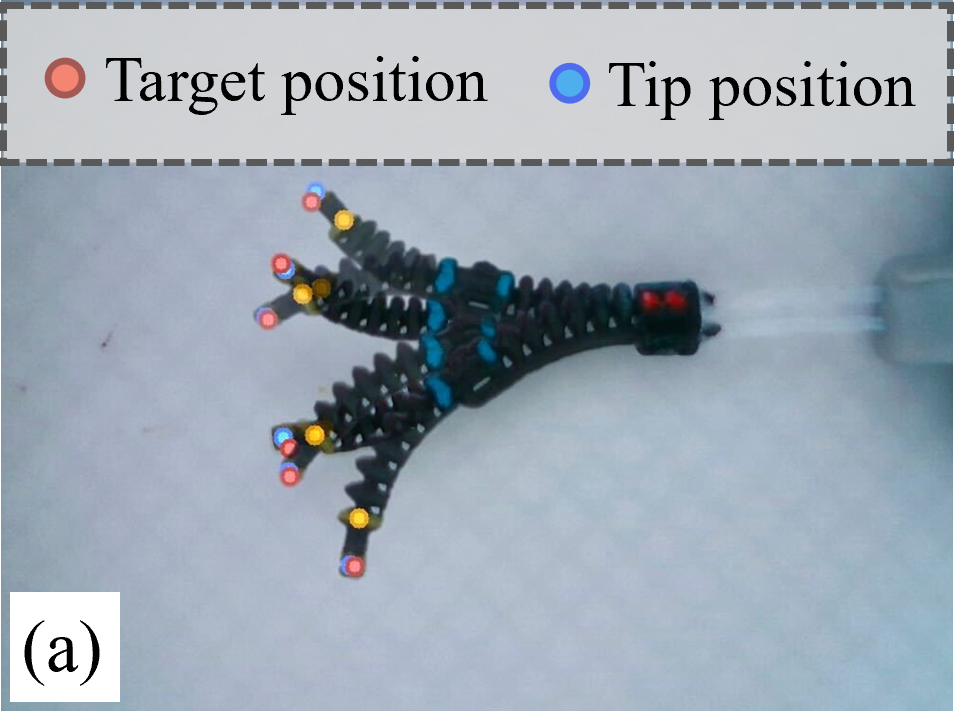}%
\label{result_a}}
\hfil
\subfloat{\includegraphics[width=1.75in]{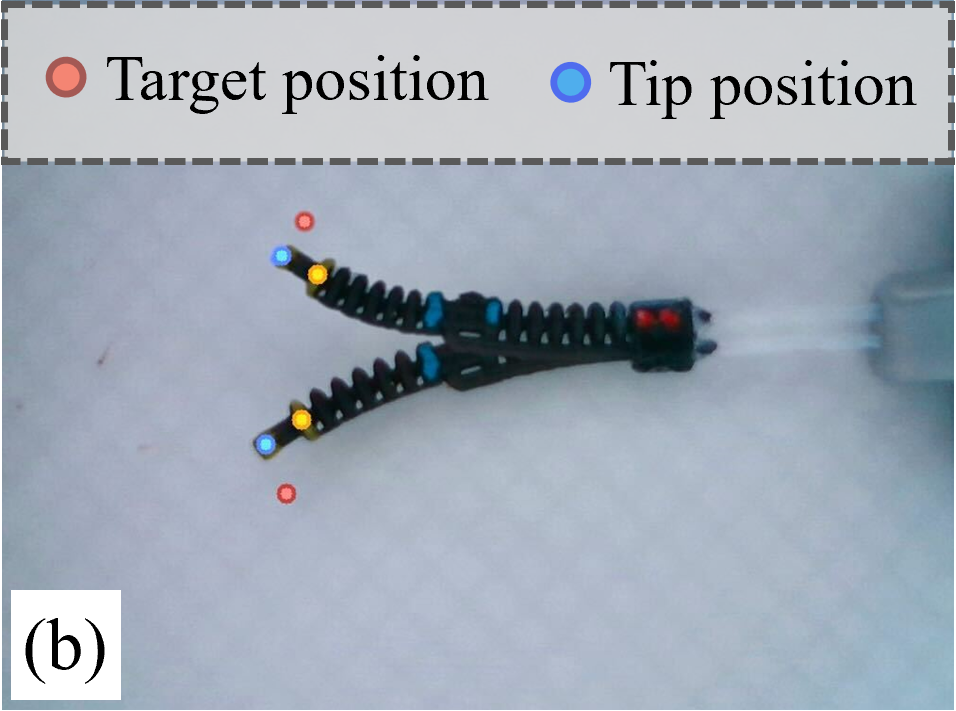}%
\label{result_b}}
\hfil
\subfloat{\includegraphics[width=1.75in]{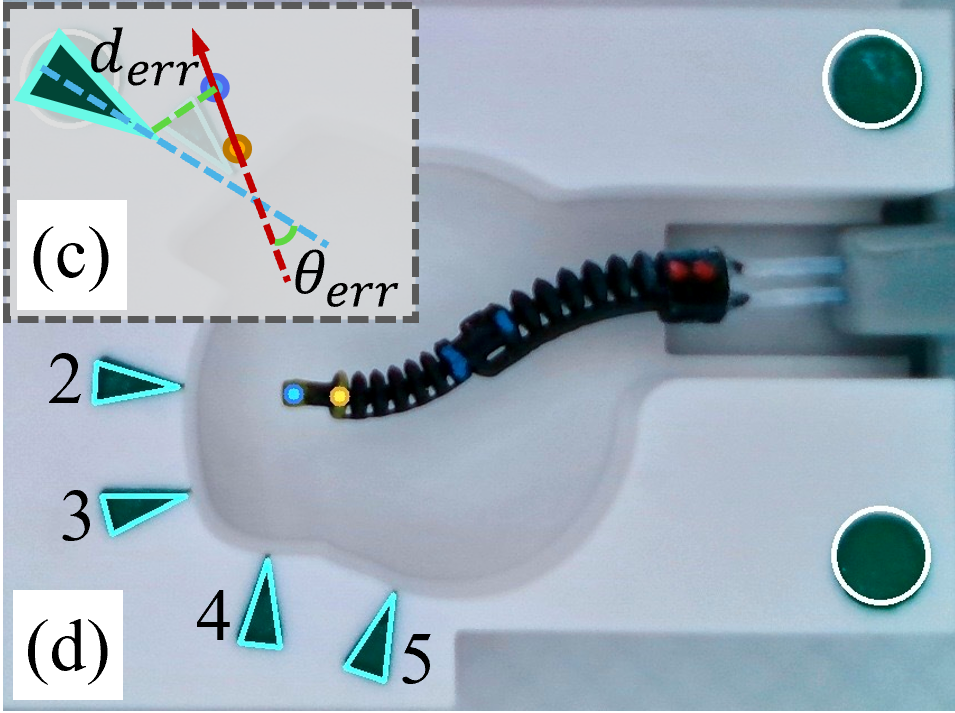}%
\label{result_c}}
\hfil
\subfloat{\includegraphics[width=1.75in]{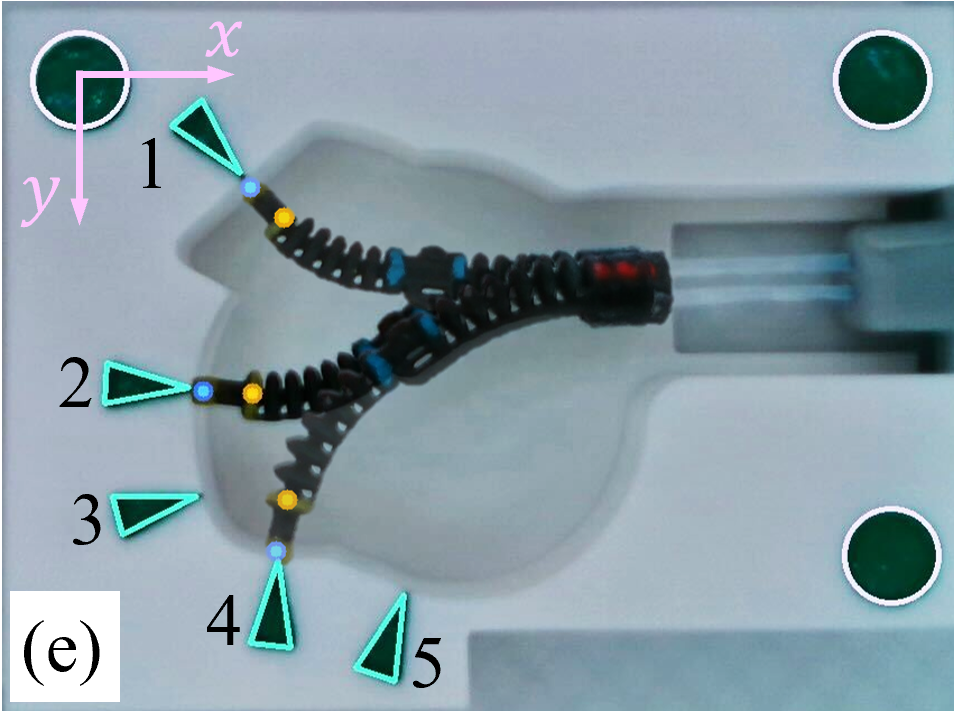}%
\label{result_d}}
\hfil
\subfloat{\includegraphics[width=3.5in]{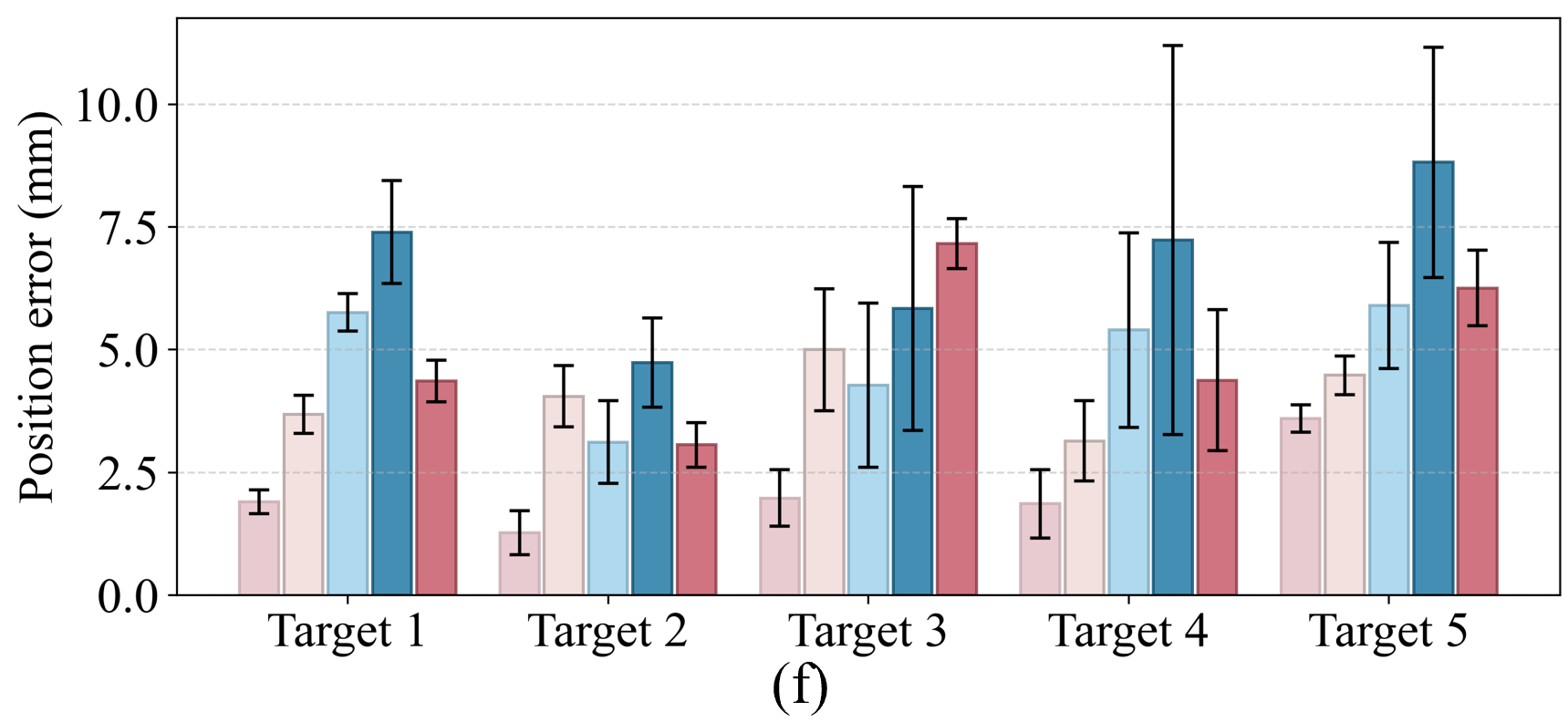}%
\label{result_3}}
\hfil
\subfloat{\includegraphics[width=3.4in]{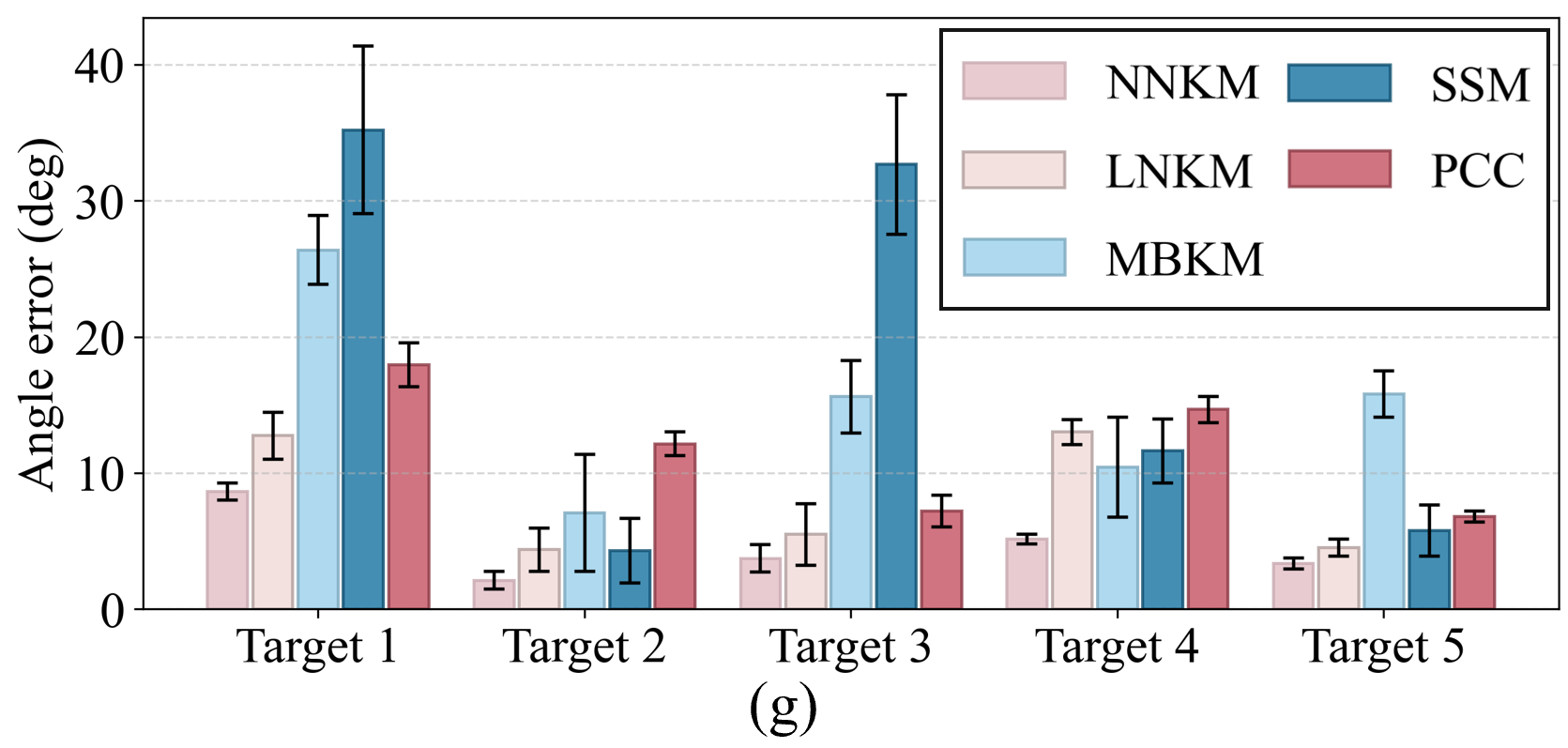}%
\label{result_f}}

\caption{Real-world experimental results. (a) Exp.~1: successful interactive position control examples using the NINK model. (b) Exp.~1: failed trial examples using other models. (c) Exp.~2: definition of position and orientation errors. (d) Exp.~2: intermediate result at Target~2 before moving the linear stage. (e) Exp.~2: Successful pose control examples. (f) Exp.~2: Distribution of position errors for each target. (g) Exp.~2: Distribution of angular errors for each target. Abbreviations: Nonlinear Input Neural Koopman + MPC(NNKM); Linear Input Neural Koopman + MPC(LNKM); Monomial Basis Koopman + MPC(MBKM); State-Space + MPC(SSM).}
\label{result}
\end{figure*}


In this experiment, we design a procedure in which the user can interactively input a desired target position to move the robot accordingly. The modeling and control state has dimension 2, corresponding to the 2D tip position. Six target positions are randomly selected within the workspace. For each trial, after the robot reaches a stable state at time $T$, the tip position is recorded by the overhead camera. The measurement error was $\pm$0.3mm. After each target is tested, the robot returns to its initial position, which is then recorded by the camera. 
Note that all tasks are performed in an open-loop scenario. To quantitatively analyze the interactive position control results, we define three evaluation metrics as follows: Average Euclidean Distance Error (AVG),
\begin{align}
\text{AVG} = \frac{1}{N} \sum_{k=1}^{N} d_k,\;\;\;k=1,2,...,N,
\end{align}
Euclidean Distance Standard Deviation (STD),
\begin{align}
\text{STD} = \sqrt{\frac{1}{N} \sum_{k=1}^{N} (d_k - \text{AVG})^2},\;\;\;k=1,2,...,N,
\end{align}
and Maximum Euclidean Distance Error (MAX),
\begin{align}
\text{MAX} = \max_{k=1,\dots,N} d_k,
\end{align}
where $N=6$ for the six target positions in total, and $d_k$ is the Euclidean distance between the target position and the catheter tip position.
For accuracy analysis, we define the Euclidean distance threshold $\sigma=p \sqrt{r_x^2+r_y^2}$, where $r_x$ and $r_y$ are the corresponding ranges in workspace, to better illustrate the performance differences of models. Specifically, we choose $p_1=2.75\%$, $p_2=5.00\%$, corresponding to thresholds $\sigma_1 = 1.5$ mm and $\sigma_2 = 3.5$ mm.

The results of the interactive position control experiment are summarized in Table \ref{tab:error_stats}. It can be seen that the NNKM achieves the best overall performance, reaching all targets within both accuracy thresholds. Fig.~\ref{result}(a) shows snapshots of all successful trials using the NNKM. LNKM demonstrates moderate performance, benefiting from its neural network-based state encoding, but its linear input assumption occasionally leads to larger errors. In contrast, MBKM and SSM perform significantly worse, reflecting the limitations of manually selected basis functions and linear assumptions, respectively. The poor modeling of these two methods also leads to unstable MPC behavior, causing the robot tip to oscillate and deviate from the target positions. Fig.~\ref{result}(b) shows snapshots of trials that did not meet the accuracy criteria.

\subsection{Experiment 2: Pose Control in an Atrium Model}

\begin{table}[t]
\vspace{2mm} 
\centering
\caption{Results of Experiment 2: Overall position (unit: mm) and orientation (unit: deg) errors, and average time cost.}
\label{tab:error_stats_2}
\begin{tabular}{lccccc}
\toprule
\textbf{Method} & \textbf{AVG}$_{\text{pos}}$ & \textbf{STD}$_{\text{pos}}$ & \textbf{AVG}$_{\text{ori}}$ & \textbf{STD}$_{\text{ori}}$ & \textbf{Time(s)}\\
\midrule
NNKM & \textbf{2.1} & \textbf{0.4} & \textbf{4.9} & \textbf{0.6} & \textbf{54.2}\\
LNKM & 4.0 & 0.7 & 8.0 & 1.4 & 58.9\\
MBKM & 4.9 & 1.2 & 15.1 & 2.9 & 71.3\\
SSM & 6.8 & 2.1 & 17.9 & 3.6 & 76.3\\
PCC & 5.0 & 0.7 & 11.8 & 1.0 & 57.1\\
\bottomrule
\end{tabular}
\end{table}


For clinically inspired evaluation, we used a 3D-printed atrium-like cavity with five triangular targets along the inner wall (Fig.~\ref{result}(c)-(d)). The length of the cavity (4.3 cm) matches the dimensions of the right atrium reported \cite{soulat2021normal}, and the entry channel (2.0 cm diameter) mimics the inferior vena cava (IVC) \cite{patil2016assessment}. The target position is defined as the vertex in contact with the atrium wall, while the orientation is determined by the angle between the midline through this vertex and the world frame. The position error and orientation error are defined as $d_{err}$ and $\theta_{err}$ in Fig.~\ref{result}(c). Detection of the colored target triangles and the reference circle is performed through contour extraction with OpenCV.

The target lies initially outside the robot’s reachable workspace. To reach it, the linear stage (see Fig.~\ref{exp}(d)) is controlled to perform a steering-like adjustment in coordination with pressure control. In this experiment, the target pose is first identified, and a corresponding pose within the workspace is determined with an offset in the $x$ direction. Then, the robot is controlled first at this intermediate pose as shown in Fig.~\ref{result}(d) when trying to reach target 2. The robot then compensates for this offset by applying a steering motion defined as $\Delta x = x_w - x_d$, where $x_w$ is the reachable $x$ coordinate in the workspace and $x_d$ is the desired target coordinate. Some successful pose control examples are shown in Fig.~\ref{result}(e). In the pose targeting experiments, each target was attempted eight times, with some trials starting from different initial states under an open-loop control scheme. The camera was used only to determine the initial states and to record the error at the final stable state.

The results of the pose control experiment in the atrium model are presented in Fig.\ref{result}(f) and Fig.\ref{result}(g). And the overall errors are summarized in Table~\ref{tab:error_stats_2}. NNKM achieves the best performance among all five control methods, with consistently lower mean and standard deviation errors in both position and orientation. 
Across all 40 trials covering the five targets, NNKM yields an average error of $2.1 \pm 0.4$mm in position and $4.9 \pm 0.6^\circ$ in orientation. LNKM exhibits moderate performance degradation compared with NNKM, with larger errors ($4.0 \pm 0.7$~mm and $8.0 \pm 1.4^\circ$) due to the simplified linear input representation, which limits its ability to capture complex actuation dynamics. In contrast, MBKM and SSM show notably larger standard deviations, particularly at targets 3, 4, and 5. Because of their weaker modeling accuracy, it is difficult for the MPC to reliably predict the robot’s motion and generate effective control inputs. 

Among baseline methods, although the PCC Jacobian control exhibits larger mean errors than NNKM and LNKM, it shows lower standard deviation than MBKM and SSM, reflecting the inherent structural stability of the PCC formulation and the use of a DLS pseudo-inverse to mitigate solution redundancy. However, the PCC approach still suffers from its assumption that each robot segment forms a perfect arc. In practice, due to material properties and fabrication imperfections, the robot experiences unintended stretching and twisting during motion, leading to inaccuracies in this model-based method.
It is also worth noting that NNKM achieves the best efficiency: on average, the time required to reach the final stable state is 8\% shorter than LNKM, 24\% shorter than MBKM, 29\% shorter than SSM, and 5\% shorter than PCC. This improvement highlights the potential of NNKM for safer and more efficient future applications in real surgical environments.

\section{Conclusion and future work}
In this study, we developed a neural Koopman operator framework combined with model predictive control to achieve accurate modeling and both position and pose control of a pneumatic soft robotic catheter. By leveraging Koopman operator theory, neural networks, and optimal control strategies, the proposed approach enables accurate state prediction and effective control in clinically inspired environments. Experimental results demonstrate that our framework outperforms existing methods in both modeling and control accuracy. To further assess applicability, we will study the generalizability of NNKM across different PSRC designs and evaluate the need for model retraining over time. We also plan to extend the framework to full-shape modeling and motion planning in constrained environments. A systematic comparison with fully nonlinear learned dynamic models \cite{thuruthel2017learning}, \cite{gillespie2018learning} remains an important future direction. In addition, catheter–tissue force estimation will be incorporated as a constraint during navigation in biological environments.These advancements will further improve the control capability of PSRCs and ensure safer operations in clinical procedures.

\bibliographystyle{IEEEtran}
\bibliography{icra.bib}

@article{bruder2020data,
  title={Data-driven control of soft robots using Koopman operator theory},
  author={Bruder, Daniel and Fu, Xun and Gillespie, R Brent and Remy, C David and Vasudevan, Ram},
  journal={IEEE transactions on robotics},
  volume={37},
  number={3},
  pages={948--961},
  year={2020},
  publisher={IEEE}
}

@article{xiao2022deep,
  title={Deep neural networks with Koopman operators for modeling and control of autonomous vehicles},
  author={Xiao, Yongqian and Zhang, Xinglong and Xu, Xin and Liu, Xueqing and Liu, Jiahang},
  journal={IEEE transactions on intelligent vehicles},
  volume={8},
  number={1},
  pages={135--146},
  year={2022},
  publisher={IEEE}
}

@article{budivsic2012applied,
  title={Applied koopmanism},
  author={Budi{\v{s}}i{\'c}, Marko and Mohr, Ryan and Mezi{\'c}, Igor},
  journal={Chaos: An Interdisciplinary Journal of Nonlinear Science},
  volume={22},
  number={4},
  year={2012},
  publisher={AIP Publishing}
}

@article{shi2022deep,
  title={Deep Koopman operator with control for nonlinear systems},
  author={Shi, Haojie and Meng, Max Q-H},
  journal={IEEE Robotics and Automation Letters},
  volume={7},
  number={3},
  pages={7700--7707},
  year={2022},
  publisher={IEEE}
}

@article{han2023robust,
  title={Robust learning and control of time-delay nonlinear systems with deep recurrent Koopman operators},
  author={Han, Minghao and Li, Zhaojian and Yin, Xiang and Yin, Xunyuan},
  journal={IEEE Transactions on Industrial Informatics},
  volume={20},
  number={3},
  pages={4675--4684},
  year={2023},
  publisher={IEEE}
}

@article{schwenzer2021review,
  title={Review on model predictive control: An engineering perspective},
  author={Schwenzer, Max and Ay, Muzaffer and Bergs, Thomas and Abel, Dirk},
  journal={The International Journal of Advanced Manufacturing Technology},
  volume={117},
  number={5},
  pages={1327--1349},
  year={2021},
  publisher={Springer}
}

@article{mauroy2019koopman,
  title={Koopman-based lifting techniques for nonlinear systems identification},
  author={Mauroy, Alexandre and Goncalves, Jorge},
  journal={IEEE Transactions on Automatic Control},
  volume={65},
  number={6},
  pages={2550--2565},
  year={2019},
  publisher={IEEE}
}

@article{webster2010design,
  title={Design and kinematic modeling of constant curvature continuum robots: A review},
  author={Webster III, Robert J and Jones, Bryan A},
  journal={The International Journal of Robotics Research},
  volume={29},
  number={13},
  pages={1661--1683},
  year={2010},
  publisher={SAGE Publications Sage UK: London, England}
}

@inproceedings{barnes2024bedside,
  title={Bedside Admittance Control of a Dual-Segment Soft Robot for Catheter-Based Interventions},
  author={Barnes, Noah and Jiang, Shaopeng and Di, Lingyun and Qu, Hannah and Janowski, Miroslaw and Berul, Charles I and Colton, Adira and Young, Olivia and Sochol, Ryan D and Brown, Jeremy D and others},
  booktitle={2024 46th Annual International Conference of the IEEE Engineering in Medicine and Biology Society (EMBC)},
  pages={1--7},
  year={2024},
  organization={IEEE}
}

@article{lai2021verticalized,
  title={Verticalized-tip trajectory tracking of a 3D-printable soft continuum robot: Enabling surgical blood suction automation},
  author={Lai, Jiewen and Huang, Kaicheng and Lu, Bo and Zhao, Qingxiang and Chu, Henry K},
  journal={IEEE/ASME transactions on mechatronics},
  volume={27},
  number={3},
  pages={1545--1556},
  year={2021},
  publisher={IEEE}
}

@inproceedings{luo2024efficient,
  title={Efficient RRT*-based safety-constrained motion planning for continuum robots in dynamic environments},
  author={Luo, Peiyu and Yao, Shilong and Yue, Yiyao and Wang, Jiankun and Yan, Hong and Meng, Max Q-H},
  booktitle={2024 IEEE International Conference on Robotics and Automation (ICRA)},
  pages={9328--9334},
  year={2024},
  organization={IEEE}
}

@inproceedings{jiang2024one,
  title={One-piece 3d-printed pneumatic catheter: Dual-segment design with integrated robotics control for endovascular interventions},
  author={Jiang, Shaopeng and Di, Lingyun and Barnes, Noah and Qu, Hannah and Young, Olivia and Brown, Jeremy D and Sochol, Ryan and Krieger, Axel},
  booktitle={2024 IEEE 7th International Conference on Soft Robotics (RoboSoft)},
  pages={832--838},
  year={2024},
  organization={IEEE}
}

@article{della2023model,
  title={Model-based control of soft robots: A survey of the state of the art and open challenges},
  author={Della Santina, Cosimo and Duriez, Christian and Rus, Daniela},
  journal={IEEE Control Systems Magazine},
  volume={43},
  number={3},
  pages={30--65},
  year={2023},
  publisher={IEEE}
}

@inproceedings{bruder2019nonlinear,
  title={Nonlinear system identification of soft robot dynamics using koopman operator theory},
  author={Bruder, Daniel and Remy, C David and Vasudevan, Ram},
  booktitle={2019 International Conference on Robotics and Automation (ICRA)},
  pages={6244--6250},
  year={2019},
  organization={IEEE}
}

@article{soulat2021normal,
  title={Normal values of right atrial size and function according to age, sex, and ethnicity: results of the world alliance societies of echocardiography study},
  author={Soulat-Dufour, Laurie and Addetia, Karima and Miyoshi, Tatsuya and Citro, Rodolfo and Daimon, Masao and Fajardo, Pedro Gutierrez and Kasliwal, Ravi R and Kirkpatrick, James N and Monaghan, Mark J and Muraru, Denisa and others},
  journal={Journal of the American Society of Echocardiography},
  volume={34},
  number={3},
  pages={286--300},
  year={2021},
  publisher={Elsevier}
}

@article{patil2016assessment,
  title={Assessment of inferior vena cava diameter by echocardiography in normal Indian population: A prospective observational study},
  author={Patil, Shivanand and Jadhav, Santosh and Shetty, Natraj and Kharge, Jayashree and Puttegowda, Beeresha and Ramalingam, Rangraj and Cholenahally, Manjunath Nanjappa},
  journal={Indian heart journal},
  volume={68},
  pages={S26--S30},
  year={2016},
  publisher={Elsevier}
}

@article{konda2025robotically,
  title={Robotically steerable guidewires—Current trends and future directions},
  author={Konda, Revanth and Brumfiel, Timothy A and Bercu, Zachary L and Grossberg, Jonathan A and Desai, Jaydev P},
  journal={Science Robotics},
  volume={10},
  number={105},
  pages={eadt7461},
  year={2025},
  publisher={American Association for the Advancement of Science}
}

@article{wijesurendra2019mechanisms,
  title={Mechanisms of atrial fibrillation},
  author={Wijesurendra, Rohan S and Casadei, Barbara},
  journal={Heart},
  volume={105},
  number={24},
  pages={1860--1867},
  year={2019},
  publisher={BMJ Publishing Group Ltd and British Cardiovascular Society}
}

@article{bassil2020robotics,
  title={Robotics for catheter ablation of cardiac arrhythmias: Current technologies and practical approaches},
  author={Bassil, Guillaume and Markowitz, Steven M and Liu, Christopher F and Thomas, George and Ip, James E and Lerman, Bruce B and Cheung, Jim W},
  journal={Journal of Cardiovascular Electrophysiology},
  volume={31},
  number={3},
  pages={739--752},
  year={2020},
  publisher={Wiley Online Library}
}

@article{shi2024position,
  title={Position and orientation control for hyperelastic multisegment continuum robots},
  author={Shi, Jialei and Abad, Sara-Adela and Dai, Jian Sheng and Wurdemann, Helge A},
  journal={IEEE/ASME Transactions on Mechatronics},
  volume={29},
  number={2},
  pages={995--1006},
  year={2024},
  publisher={IEEE}
}

@article{ferrentino2023finite,
  title={Finite element analysis-based soft robotic modeling: Simulating a soft actuator in sofa},
  author={Ferrentino, Pasquale and Roels, Ellen and Brancart, Joost and Terryn, Seppe and Van Assche, Guy and Vanderborght, Bram},
  journal={IEEE robotics \& automation magazine},
  volume={31},
  number={3},
  pages={97--105},
  year={2023},
  publisher={IEEE}
}

@article{laschi2023learning,
  title={Learning-based control strategies for soft robots: Theory, achievements, and future challenges},
  author={Laschi, Cecilia and Thuruthel, Thomas George and Lida, Fumiya and Merzouki, Rochdi and Falotico, Egidio},
  journal={IEEE Control Systems Magazine},
  volume={43},
  number={3},
  pages={100--113},
  year={2023},
  publisher={IEEE}
}

@article{hao2023inverse,
  title={Inverse kinematic modeling of the tendon-actuated medical continuum manipulator based on a lightweight timing input neural network},
  author={Hao, Jianxiong and Duan, Jinyu and Wang, Kaifeng and Hu, Chengzhi and Shi, Chaoyang},
  journal={IEEE Transactions on Medical Robotics and Bionics},
  volume={5},
  number={4},
  pages={916--928},
  year={2023},
  publisher={IEEE}
}

@article{yao2025adaptive,
  title={Adaptive Load-Dependent Sim2Real Framework for Path Tracking Toward Tendon-Driven Continuum Robots},
  author={Yao, Shilong and Luo, Peiyu and Yue, Yiyao and Chen, Yuhan and Yan, Hong and Meng, Max Q-H},
  journal={IEEE/ASME Transactions on Mechatronics},
  year={2025},
  publisher={IEEE}
}

@article{fang2022efficient,
  title={Efficient Jacobian-based inverse kinematics with sim-to-real transfer of soft robots by learning},
  author={Fang, Guoxin and Tian, Yingjun and Yang, Zhi-Xin and Geraedts, Jo MP and Wang, Charlie CL},
  journal={IEEE/ASME Transactions on Mechatronics},
  volume={27},
  number={6},
  pages={5296--5306},
  year={2022},
  publisher={IEEE}
}

@inproceedings{komeno2022deep,
  title={Deep koopman with control: Spectral analysis of soft robot dynamics},
  author={Komeno, Naoto and Michael, Brendan and K{\"u}chler, Katharina and Anarossi, Edgar and Matsubara, Takamitsu},
  booktitle={2022 61st Annual Conference of the Society of Instrument and Control Engineers (SICE)},
  pages={333--340},
  year={2022},
  organization={IEEE}
}

@article{ferguson2009catheter,
  title={Catheter ablation of atrial fibrillation without fluoroscopy using intracardiac echocardiography and electroanatomic mapping},
  author={Ferguson, John D and Helms, Adam and Mangrum, J Michael and Mahapatra, Srijoy and Mason, Pamela and Bilchick, Ken and McDaniel, George and Wiggins, David and DiMarco, John P},
  journal={Circulation: Arrhythmia and Electrophysiology},
  volume={2},
  number={6},
  pages={611--619},
  year={2009},
  publisher={Lippincott Williams \& Wilkins}
}

@article{enriquez2024feasibility,
  title={Feasibility, efficacy, and safety of fluoroless ablation of VT in patients with structural heart disease},
  author={Enriquez, Andres and Sadek, Mouhannad and Hanson, Matthew and Yang, Jaejoon and Matos, Carlos D and Neira, Victor and Marchlinski, Francis and Miranda-Arboleda, Andres and Orellana-C{\'a}ceres, Juan-Jos{\'e} and Alviz, Isabella and others},
  journal={Clinical Electrophysiology},
  volume={10},
  number={7\_Part\_1},
  pages={1287--1300},
  year={2024},
  publisher={American College of Cardiology Foundation Washington DC}
}

@article{prolivc2022conventional,
  title={Conventional fluoroscopy-guided versus zero-fluoroscopy catheter ablation of supraventricular tachycardias},
  author={Proli{\v{c}} Kalin{\v{s}}ek, Tine and {\v{S}}orli, Jernej and Jan, Matev{\v{z}} and {\v{S}}inkovec, Matja{\v{z}} and Antoli{\v{c}}, Bor and Klemen, Luka and {\v{Z}}i{\v{z}}ek, David and Pernat, Andrej},
  journal={BMC Cardiovascular Disorders},
  volume={22},
  number={1},
  pages={98},
  year={2022},
  publisher={Springer}
}

@article{thamo2022data,
  title={Data-driven steering of concentric tube robots in unknown environments via dynamic mode decomposition},
  author={Thamo, Balint and Hanley, David and Dhaliwal, Kevin and Khadem, Mohsen},
  journal={IEEE Robotics and Automation Letters},
  volume={8},
  number={2},
  pages={856--863},
  year={2022},
  publisher={IEEE}
}

@article{thuruthel2017learning,
  title={Learning dynamic models for open loop predictive control of soft robotic manipulators},
  author={Thuruthel, Thomas George and Falotico, Egidio and Renda, Federico and Laschi, Cecilia},
  journal={Bioinspiration \& biomimetics},
  volume={12},
  number={6},
  pages={066003},
  year={2017},
  publisher={IOP Publishing}
}

@inproceedings{gillespie2018learning,
  title={Learning nonlinear dynamic models of soft robots for model predictive control with neural networks},
  author={Gillespie, Morgan T and Best, Charles M and Townsend, Eric C and Wingate, David and Killpack, Marc D},
  booktitle={2018 IEEE International Conference on Soft Robotics (RoboSoft)},
  pages={39--45},
  year={2018},
  organization={IEEE}
}

\end{document}